\documentclass{article}

     \PassOptionsToPackage{numbers, compress}{natbib}
\usepackage[preprint]{neurips_2020}

\usepackage[utf8]{inputenc} % allow utf-8 input
\usepackage[T1]{fontenc}    % use 8-bit T1 fonts
\usepackage{hyperref}       % hyperlinks
\usepackage{url}            % simple URL typesetting
\usepackage{booktabs}       % professional-quality tables
\usepackage{amsfonts}       % blackboard math symbols
\usepackage{nicefrac}       % compact symbols for 1/2, etc.
\usepackage{microtype}      % microtypography
\usepackage{amsmath}
\usepackage{graphicx}
\usepackage{subcaption}
\usepackage{dsfont}
\usepackage{caption}
\usepackage{relsize}

\title{Trustworthy Convolutional Neural Networks: \\A Gradient Penalized-based Approach}
\author{%
  Nicholas Halliwell \\
  Inria, Sophia Antipolis, France \\
  \texttt{nicholas.halliwell@inria.fr} \\
  \And
  Freddy Lecue \\
  CortAIx, Thales, Montreal, Canada \\
  Inria, Sophia Antipolis, France \\
  \texttt{freddy.lecue@inria.fr} \\
}

\begin{document}

\maketitle

\begin{abstract}
Convolutional neural networks (CNNs) are commonly used for image classification. Saliency methods are examples of approaches that can be used to interpret CNNs post hoc, identifying the most relevant pixels for a prediction following the gradients flow. Even though CNNs can correctly classify images, the underlying saliency maps could be erroneous in many cases. This can result in skepticism as to the validity of the model or its interpretation. We propose a novel approach for training trustworthy CNNs by penalizing parameter choices that result in inaccurate saliency maps generated during training. We add a penalty term for inaccurate saliency maps produced when the predicted label is correct, a penalty term for accurate saliency maps produced when the predicted label is incorrect, and a regularization term penalizing overly confident saliency maps. Experiments show increased classification performance, user engagement, and trust.
\end{abstract}
%ABSTRACT MUST BE ONE PARAGRAPH ONLY
%NO VERTICAL LINES ON TABLES
\section{Introduction}
Convolutional neural networks (CNNs) are used in computer vision for tasks such as object detection~\citep{DBLP:journals/corr/abs-1804-02767,DBLP:journals/corr/RedmonDGF15, DBLP:conf/cvpr/RedmonF17}, image classification~\cite{DBLP:conf/nips/KrizhevskySH12}, and visual question answering~\citep{DBLP:conf/iccv/AntolALMBZP15, DBLP:conf/nips/CadeneDBCP19}. The success of these models has created a need for model transparency. Due to their complex structure, CNNs are often difficult to interpret. Features learned by CNNs can be visualized, but understanding the meaning of these hidden representations can be difficult for non-experts.

Indeed there are many approaches to interpreting the output of machine learning models~\citep{DBLP:conf/aaai/Ribeiro0G18,DBLP:conf/kdd/Ribeiro0G16,DBLP:conf/nips/LundbergL17,DBLP:conf/iclr/AnconaCO018, DBLP:journals/corr/MontavonBBSM15, DBLP:conf/icml/KimWGCWVS18, DBLP:conf/nips/HookerEKK19, DBLP:conf/nips/ChenLTBRS19}. Saliency methods offer an intuitive way to understand what a CNN has learned. These algorithms provide post hoc interpretations by highlighting a set of pixels or super pixels in the input image that are most relevant for a prediction. Small changes to these highlighted pixels results in the biggest change in predicted score. One of the first saliency methods, Gradients (sometimes called Vanilla Gradients) compute the gradient of the class score with respect to the input image~\cite{DBLP:journals/corr/SimonyanVZ13}. Guided BackPropagation~\cite{DBLP:journals/corr/SpringenbergDBR14} imputes the gradient of layers using ReLUs, backpropagating only positive gradients. Class Activation Mapping (CAM) uses a specific architecture for CNNs to discriminate regions of the input~\cite{DBLP:conf/cvpr/ZhouKLOT16}. Grad-CAM generalizes CAM to any CNN architecture~\cite{DBLP:journals/corr/SelvarajuDVCPB16}. There exists many different saliency methods~\citep{DBLP:conf/icml/ShrikumarGK17, DBLP:journals/corr/ShrikumarGSK16, DBLP:conf/eccv/ZeilerF14, DBLP:journals/corr/SmilkovTKVW17, DBLP:conf/icml/SundararajanTY17, DBLP:journals/pr/MontavonLBSM17} to visually interpret CNN predictions. There is existing work showing not all of these methods are robust~\cite{DBLP:conf/nips/AdebayoGMGHK18}, and that saliency maps can be manipulated~\cite{DBLP:conf/nips/DombrowskiAAAMK19}. HINT~\cite{DBLP:conf/iccv/SelvarajuLSJGHB19} encourages deep neural networks to be sensitive to the same input regions as humans. The authors of~\cite{DBLP:journals/corr/abs-1909-13584} add a penalty term for explanation predictions far away from their respective ground truth. For future versions, potential baselines include~\citep{DBLP:conf/iclr/ZintgrafCAW17,DBLP:conf/iclr/ChangCGD19,DBLP:conf/iclr/KindermansSAMEK18}

On the task of image classification, saliency methods allow the user to visually inspect the highlighted regions of the image. The overlap between the saliency map and the object of interest can be easily and quickly compared by users. This forms an intuitive way to interpret what a model has learned. Saliency methods however can attribute relevant pixels outside the object of interest, producing what we term an inaccurate saliency map. Indeed any of the saliency methods mentioned above can encounter some form of this issue. This indicates the model has chosen a poor set of parameters.\footnote{We recognize it is also possible the saliency method is not robust. For this paper we focus on model induced errors.} On the contrary, an accurate saliency map attributes pixel values over the object of interest. 

\begin{figure*}[h]
	\centering
	\begin{subfigure}{0.27\textwidth}
		\includegraphics[width=\textwidth]{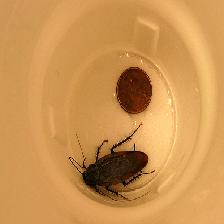}
		\caption{Input Image: Cockroach}
		\label{fig:example-input}
	\end{subfigure}
	\begin{subfigure}{0.27\textwidth}
		\includegraphics[width=\textwidth]{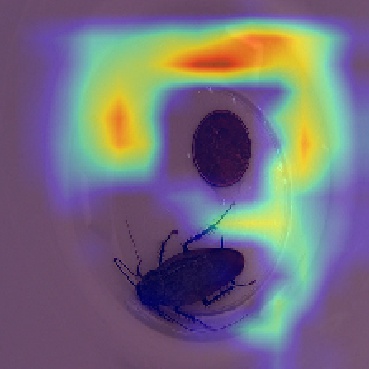}
		\caption{VGG-16: Tick}
		\label{fig:example-base}
	\end{subfigure}
	\begin{subfigure}{0.27\textwidth}
	\includegraphics[width=\textwidth]{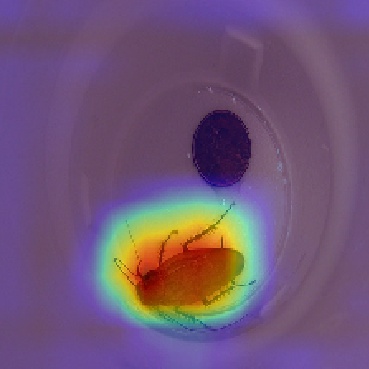}
	\caption{Proposed: Cockroach}
	\label{fig:example-repair}
	\end{subfigure}
	\caption{VGG-16 incorrectly classifies the input image, and the post hoc saliency map is inaccurate.The trustworthy CNN correctly classifies the image and produces an accurate saliency map.}
	\label{fig:example-fig}
\end{figure*}

Convolutional neural networks are most commonly optimized by minimizing the cross entropy between the target distribution and the predicted target distribution. This loss function is unaware that the choice of parameters giving a correct prediction may result in inaccurate saliency maps shown to the user. Conversely, parameters may be learned that give an incorrect prediction but result in a visually accurate saliency map. Figure~\ref{fig:example-base} shows an example where the predicted label is incorrect, and the saliency map is inaccurate. Figure~\ref{fig:example-repair} shows an example where the predicted label is correct, and the saliency map is accurate.

We propose a loss function that unifies traditional loss functions with post hoc interpretation methods. This function includes a penalty term for inaccurate saliency maps generated when the predicted class is correct, a penalty term for accurate saliency maps generated when the prediction is incorrect, and a penalty term for overly confident saliency maps. Indeed this involves computing predicted saliency maps on the forward pass during training. We demonstrate these penalty terms can be added to the existing loss of a pre-trained model to continue training, or used in a transfer learning framework to improve post hoc saliency maps. 

\section{Problem Setting}
\subsection{Prior Work}
Consider a convolutional neural network $f$ that classifies an image $x_{i}$ into one of $c \in C$ classes. Let $y^{c}$ be the true label $c$ of an image $x_{i}$. Let $\hat{y}$ be $x_{i}$'s predicted label from $f$, or equivalently $\hat{y} = f(x_{i})_{c}$. The cross entropy is thus given by

\begin{align}
CE(y, \hat{y}) = \frac{1}{n} \sum\limits_{i=1}^{n}\sum\limits_{c=1}^{C} \mathds{-1}[y^{i} = c] \,log(f(x_{i})_{c})
\label{eq:ce}
\end{align}

Two saliency methods of particular interest are Grad-CAM and Guided Grad-CAM~\cite{DBLP:journals/corr/SelvarajuDVCPB16}. Let $A$ be the activation maps of some convolutional layer, a given activation map denoted $A^{k}$. To produce a saliency map for some $x_{i}$, Grad-CAM computes the gradients of the target class with respect to layer $l$'s activation maps. Global average pooling is performed on the gradients to serve as weights for each activation map. These weights, denoted $\alpha_{k}^{c}$ in equation~\ref{eq:alpha} represent the importance of a given activation map $k$ for some target class $c \in C$.

\begin{align}
\alpha_{k}^{c} = \frac{1}{W \times H} \sum\limits_{i} \sum\limits_{j} \frac{ \partial y^{c}}{\partial A^{k}_{ij}}
\label{eq:alpha}
\end{align}

After computing a weighted combination of the activation maps, the resulting output is passed through a ReLU. Here a ReLU is used to focus only on the features that have a positive influence on the true class $c$, i.e. pixels whose intensity should be increased in order to increase $y^{c}$~\cite{DBLP:journals/corr/SelvarajuDVCPB16}. The saliency map output by Grad-CAM is thus given by %$L_{Grad-CAM}^{c}$ with dimensions equal to the original the activation map $A$. 

\begin{align}
L^{c}_{(Grad-CAM)} = ReLU \left(\sum_{k} \alpha_{k}^{c} A^{k} \right)
\label{eq:grad-cam}
\end{align}

Guided Grad-CAM~\cite{DBLP:journals/corr/SelvarajuDVCPB16} combines the output from Grad-CAM (equation~\ref{eq:grad-cam}) with the output from Guided Backpropagation~\cite{DBLP:journals/corr/SpringenbergDBR14} through element-wise multiplication. This is done for two reasons; First, Guided Backpropagation alone is not class discriminative. Second, Grad-CAM fails to produce high resolution (fine-grained) visualizations. Merging the two saliency methods produces saliency maps that are both high resolution and class discriminative~\cite{DBLP:journals/corr/SelvarajuDVCPB16}.

\subsection{Shortcomings}
Despite state-of-the-art classification performance achieved by convolutional neural networks, loss minimizing parameters may result in saliency maps that do not highlight relevant pixels over the object of interest. Indeed the saliency map is dependent on the learned parameters. Parameters however are learned without knowing if the resulting saliency maps are visually accurate. Additionally, showing an inaccurate saliency map to a practitioner does not provide insight on how to change model parameters to correctly highlight pixels over the object of interest. 

\begin{figure*}[t]
	\centering
	\begin{subfigure}{0.19\textwidth}
		\includegraphics[width=\textwidth]{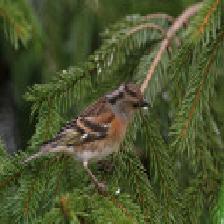}
		\caption{True class:\\Brambling}
	\end{subfigure}
	\begin{subfigure}{0.19\textwidth}
		\includegraphics[width=\textwidth]{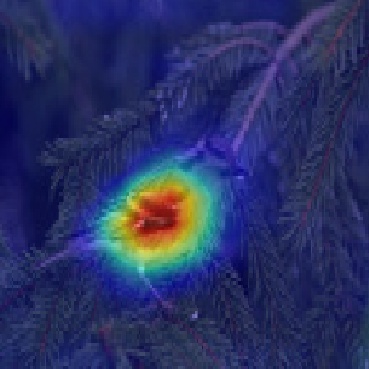}
		\caption{Predicted class:\\Brambling}
		\label{fig:case-1}
	\end{subfigure}
	\begin{subfigure}{0.19\textwidth}
		\includegraphics[width=\textwidth]{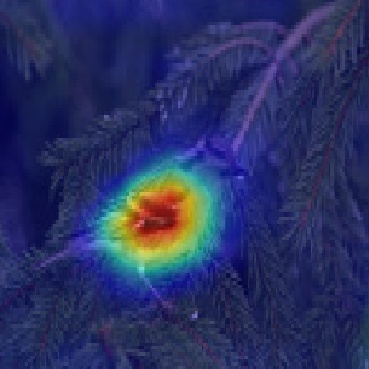}
		\caption{Predicted class:\\Brambling}
	\end{subfigure}
	\begin{subfigure}{0.19\textwidth}
		\includegraphics[width=\textwidth]{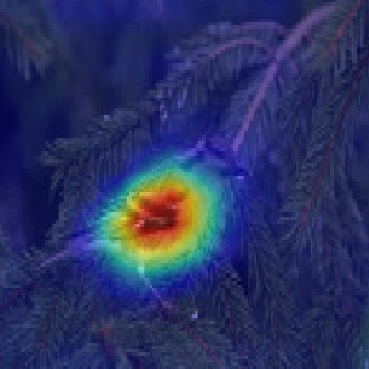}
		\caption{Predicted class:\\Brambling}
	\end{subfigure}
	\begin{subfigure}{0.19\textwidth}
		\includegraphics[width=\textwidth]{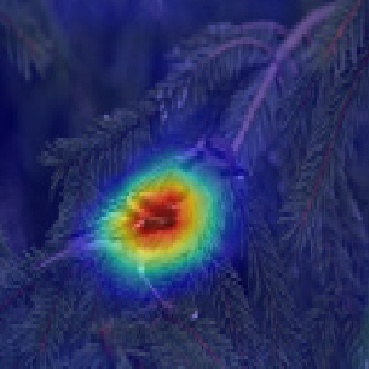}
		\caption{Predicted class:\\Brambling}
	\end{subfigure}
	\begin{subfigure}{0.19\textwidth}
		\includegraphics[width=\textwidth]{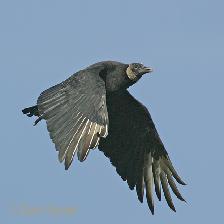}
		\caption{True class:\\Vulture}
	\end{subfigure}
	\begin{subfigure}{0.19\textwidth}
		\includegraphics[width=\textwidth]{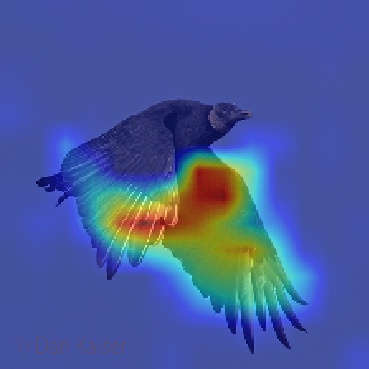}
		\caption{Predicted class:\\Kite}
		\label{fig:case-2}
	\end{subfigure}
	\begin{subfigure}{0.19\textwidth}
		\includegraphics[width=\textwidth]{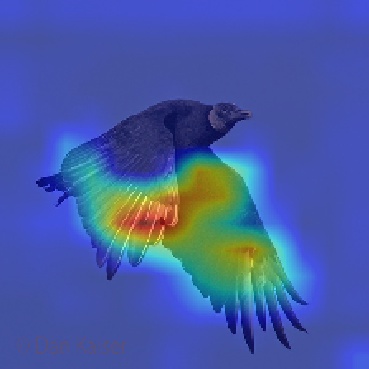}
		\caption{Predicted class:\\Vulture}
	\end{subfigure}
	\begin{subfigure}{0.19\textwidth}
		\includegraphics[width=\textwidth]{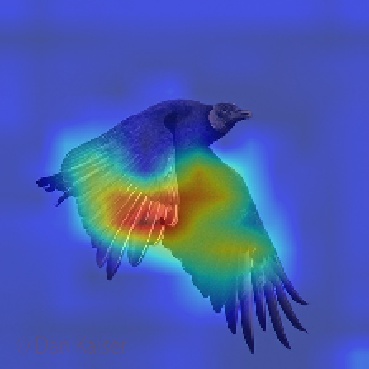}
		\caption{Predicted class:\\Vulture}
	\end{subfigure}
	\begin{subfigure}{0.19\textwidth}
		\includegraphics[width=\textwidth]{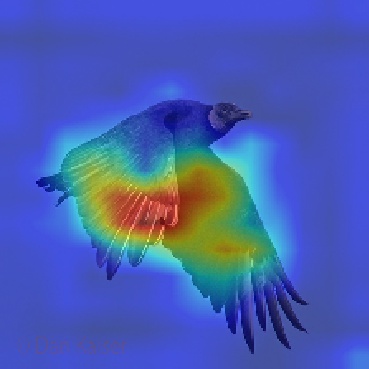}
		\caption{Predicted class:\\Vulture}
	\end{subfigure}
	\begin{subfigure}{0.19\textwidth}
		\includegraphics[width=\textwidth]{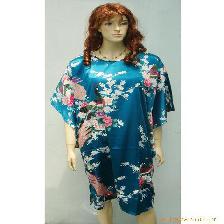}
		\caption{True class:\\Pajama}
	\end{subfigure}
	\begin{subfigure}{0.19\textwidth}
		\includegraphics[width=\textwidth]{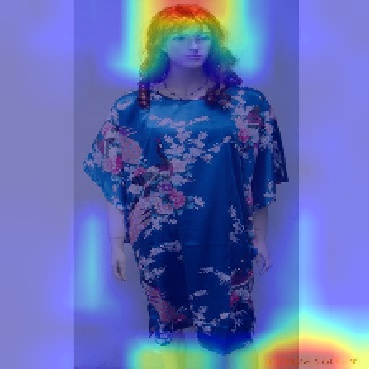}
		\caption{Predicted class:\\Pajama}
		\label{fig:case-3}
	\end{subfigure}
	\begin{subfigure}{0.19\textwidth}
		\includegraphics[width=\textwidth]{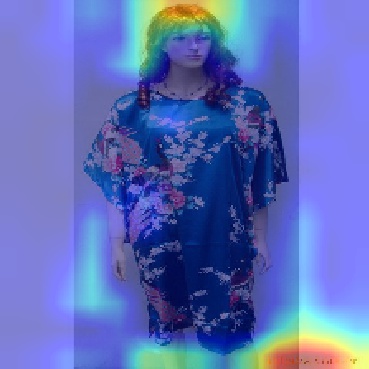}
		\caption{Predicted class:\\Pajama}
	\end{subfigure}
		\begin{subfigure}{0.19\textwidth}
		\includegraphics[width=\textwidth]{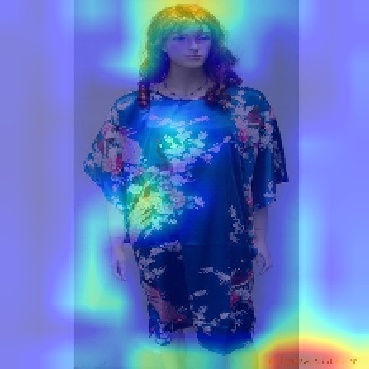}
		\caption{Predicted class:\\Pajama}
	\end{subfigure}
	\begin{subfigure}{0.19\textwidth}
		\includegraphics[width=\textwidth]{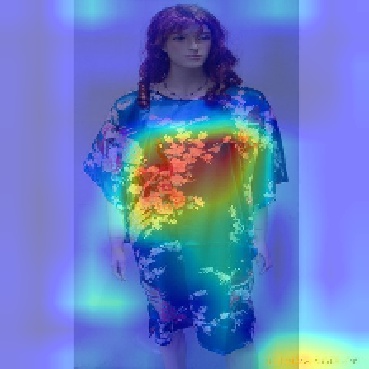}
		\caption{Predicted class:\\Pajama}
	\end{subfigure}
	\begin{subfigure}{0.19\textwidth}
		\includegraphics[width=\textwidth]{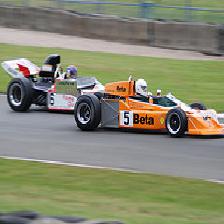}
		\caption{True class:\\Sports car}
		\caption*{\bf{Input image}}
	\end{subfigure}
	\begin{subfigure}{0.19\textwidth}
		\includegraphics[width=\textwidth]{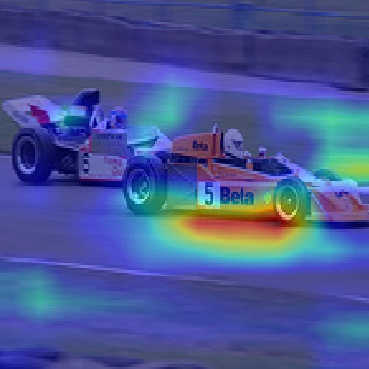}
		\caption{Predicted class:\\Racer}
		\caption*{\bf{Baseline}}
		\label{fig:case-4}
	\end{subfigure}
	\begin{subfigure}{0.19\textwidth}
		\includegraphics[width=\textwidth]{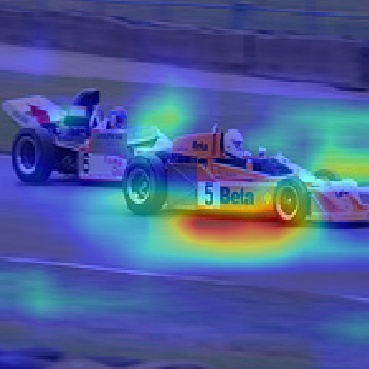}
		\caption{Predicted class:\\Sports car}
		\caption*{\bf{Proposed ($.01$)}}
	\end{subfigure}
	\begin{subfigure}{0.19\textwidth}
		\includegraphics[width=\textwidth]{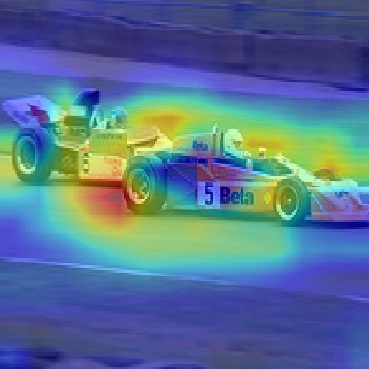}
		\caption{Predicted class:\\Sports car}
		\caption*{\bf{Proposed ($.1$)}}
	\end{subfigure}
	\begin{subfigure}{0.19\textwidth}
		\includegraphics[width=\textwidth]{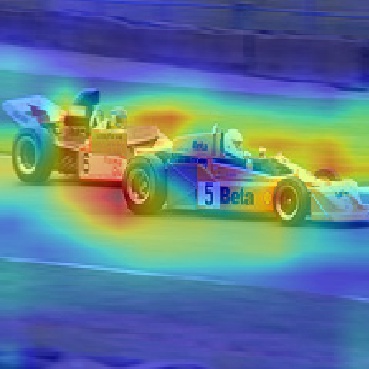}
		\caption{Predicted class:\\Sports car}
		\caption*{\bf{Proposed ($.3$)}}
	\end{subfigure}
	\caption{Input image shown with post hoc saliency maps from a VGG-16 baseline, and our proposed gradient penalized based Trustworthy CNN model shown with various learning rates.}
	\label{fig:saliency-maps}
\end{figure*}

As an example, take a pre-trained model VGG-16~\cite{DBLP:journals/corr/SimonyanZ14a}, trained on the ImageNet dataset~\cite{DBLP:journals/ijcv/RussakovskyDSKS15}. We use Grad-CAM on selected images to identify relevant pixels. Figure~\ref{fig:saliency-maps} shows four different cases that can be encountered when using saliency methods for interpretations. Figure~\ref{fig:case-1} shows the case $1$, where the predicted label is correct, and the resulting saliency map is visually accurate. Figure~\ref{fig:case-2} shows case $2$, where the predicted class is incorrect, and the resulting saliency map is correct. Figure~\ref{fig:case-3} shows case $3$, the predicted class is correct, and the resulting saliency map is inaccurate. Figure~\ref{fig:case-4} shows case $4$, where the predicted class is incorrect and the resulting saliency map is inaccurate.

Models giving inaccurate predictions (Figures~\ref{fig:case-2},~\ref{fig:case-4}) and/or inaccurate saliency maps (Figures~\ref{fig:case-3},~\ref{fig:case-4}) will cause users to lose trust in the model. Currently, convolutional neural networks are optimized ignoring how the saliency map will look post hoc. To our knowledge, no method exists to train convolutional neural networks to produce visually accurate saliency maps. We propose a loss function that penalizes inaccurate saliency maps, resulting in model parameters that produce visually accurate saliency maps post hoc, and improved classification performance, ensuring better user trust. 

\section{Trustworthy Convolutional Neural Networks}
We define a trustworthy CNN as one that produces accurate predictions and visually accurate post hoc saliency maps determined by user evaluation. %A naive approach to learn a trustworthy CNN would be to use segmented data where the object of interest is already identified, and penalize parameter choices that attribute pixels outside of the bounding box. This would require annotated data, which could be difficult if not impossible in many application domains. We propose a loss function that produces visually accurate saliency maps without the need for annotated training data. 

\subsection{Loss function}

To identify parameters that produce both accurate predictions and accurate saliency maps, constraints must to be added to the cross entropy loss. Saliency maps produced post hoc can be visually accurate while the model classifies the observation incorrectly. Additionally, visually inaccurate saliency maps can be produced while the model classifies the observation correctly. Lastly, visually inaccurate saliency maps can be observed while the model incorrectly classifies the observation. The loss function must consider the saliency maps produced from the parameter choices at each step taken by the optimizer. 

Take a saliency map $\hat{L}^{c}_{(.)_{i}}$ generated by a saliency method on the forward pass of training. We average the predicted saliency map across all dimensions. Given by equation~\ref{eq:regularizer}, we use this penalty term to gauge the confidence of the predicted saliency map. 

% The dimensions are given by $Z= W \times H$.

\begin{align}
\hat{S}_{i} = \frac{1}{W \times H} \sum\limits_{w=1}^{W} \sum\limits_{h=1}^{H} \hat{L}^{c}_{(.)_{i}}
\label{eq:regularizer}
\end{align}

Adding the constraint of overly confident saliency maps generated during training does not penalize interactions between the saliency maps and predicted labels. Further constraints are needed to account for the predicted saliency map being accurate when the predicted label is incorrect, and the predicted saliency map being inaccurate when the predicted label is correct. Equation~\ref{eq:R1} and~\ref{eq:R2} are added for the interaction between the predicted class labels and predicted saliency map. Large gradient saliency maps with corresponding incorrect predicted labels are penalized, along with small gradient saliency maps with corresponding correct predicted labels.

%is added to penalize large gradient saliency maps when the predicted label is incorrect. 
%Equation is added to penalize small gradient saliency maps when the predicted label is correct.

\begin{align}
%(y_{i}, \hat{y}_{i},S_{i}, \hat{S}_{i})
R_{1} = CE(y_{i}, \hat{y}_{i})(1 - \hat{S}_{i})
\label{eq:R1}
\end{align}

\begin{align}
%(y_{i}, \hat{y}_{i}, S_{i}, \hat{S}_{i})
R_{2} = \hat{S}_{i}(1 - CE(y_{i}, \hat{y}_{i}))
\label{eq:R2}
\end{align}

The final loss function used for all plots and tables in this work is given by equation~\ref{eq:loss}. We use a scalar $\lambda \in [0,1]$ to establish a dependence between $\hat{S}_{i}$ and the cross entropy $CE$. \footnote{We recognize $\hat{S}_{i}$ and $CE$ are not guaranteed to be between zero and one. In our experiments however, we find that the cross entropy term in equation~\ref{eq:ce} and regularization term in equation~\ref{eq:regularizer} are between zero and one when each term is divided by the number of classes $|C|$.} 
\begin{align}
\mathcal{L}(y_, \hat{y},S) = \sum_{i=1}^{n} \lambda CE(y_{i}, \hat{y}_{i}) + (1 - \lambda)S_{i} + R_{1} + R_{2}
\label{eq:loss}
\end{align}

The loss function we plan on using in future versions is given by 

\begin{align}
%(y_{i}, \hat{y}_{i},S_{i}, \hat{S}_{i})
R_{1} = CE(y_{i}, \hat{y}_{i})(1 - PWCE(S_{i},\hat{S}_{i}))
\end{align}

\begin{align}
%(y_{i}, \hat{y}_{i}, S_{i}, \hat{S}_{i})
R_{2} = PWCE(S_{i},\hat{S}_{i})(1 - CE(y_{i}, \hat{y}_{i}))
\end{align}

\begin{align}
\mathcal{L}(y_, \hat{y},S,\hat{S}
) = \sum_{i=1}^{n} \lambda CE(y_{i}, \hat{y}_{i}) + (1 - \lambda)PWCE(S_{i},\hat{S}_{i})  + R_{1} + R_{2}
\end{align}

where $PWCE$ is the pixel-wise cross entropy between the ground truth saliency map and predicted saliency map. 

\subsection{Training}
To optimize the loss proposed in equation~\ref{eq:loss}, we freeze the weights of all other layers in the network. We use stochastic gradient descent in all our experiments, although any of its variants can be used.  

Naturally, this loss function will be most effective in two settings; to update previously learned parameters of a pre-trained model, or learn parameters of a newly added layer in a transfer learning framework. Consider the following example, where a practitioner identifies a layer in a convolutional network that learns a noticeable systematic error. Our loss function allows practitioners to update the layer weights, and eliminate these errors without having to re-train the model from scratch. In the case of transfer learning, a new layer can be added and parameters can be learned that will produce accurate saliency maps post hoc.

%%%%%%%%%%%%%%%%%%%%%%%%

There are no restrictions on which saliency method can be used to produce the saliency maps $L^{c}_{(.)_{i}}$ generated during training, provided the generated output when averaged is between zero and one. Regarding choice of saliency method, some choices make more intuitive sense than others. For example, Guided Backpropagation~\cite{DBLP:journals/corr/SpringenbergDBR14} and Deconvolutions~\cite{DBLP:conf/eccv/ZeilerF14} are not class discriminative, and therefore should not be chosen.

\section{Experimental Studies}
\subsection{Transfer Learning}

One interesting application of the proposed loss function is its application to the field of transfer learning~\cite{DBLP:journals/pami/Fei-FeiFP06}. Knowledge from models trained on a specific task are applied to an entirely different domain. Some recent developments~\citep{DBLP:conf/nips/RaghuZKB19,DBLP:conf/nips/LeeSW19, DBLP:conf/nips/SongCWSS19,DBLP:conf/nips/HannekeK19, DBLP:conf/ijcai/ZhuangCLPH15}, and several surveys~\citep{DBLP:journals/tkde/PanY10, DBLP:journals/corr/abs-1911-02685} can provide further detail.

We demonstrate how the proposed loss can be used in the context of transfer learning to improve classification performance. For this experiment, we use MobilenetV2~\cite{DBLP:conf/cvpr/SandlerHZZC18} on the cats and dogs dataset~\cite{DBLP:conf/ccs/ElsonDHS07}, found in Tensorflow~\cite{DBLP:conf/osdi/AbadiBCCDDDGIIK16}. The task is to classify whether an image contains a cat or dog, using the knowledge learned from the ImageNet dataset. We place one convolutional layer after the last convolutional layer in the MobilenetV2 network, closest to the softmax layer. This additional convolutional layer consists of $256$ filters, using a $1\times1$ kernel with a stride of $1$. We then remove all layers after, and add a softmax layer. All other layer weights are frozen. The baseline model uses the cross entropy loss given by equation~\ref{eq:ce}. We compare this against two trustworthy CNN models trained using equation~\ref{eq:loss}. The first uses Grad-CAM to generate saliency maps during training, the second uses Guided Grad-CAM. We train all models for $50$ epochs using a batch size of $32$. We compare post hoc saliency maps using the structured similarity index (SSIM) given by equation~\ref{eq:ssim}, to compare relative to the baseline model. We fix the learning rate to $.01$ and set $\lambda = .9$.
\begin{align}
    SSIM(x,y) = \frac{(2\mu_{x}\mu_{y} + c_{1})(2\sigma_{xy} + c_{2})}{(\mu^{2}_{x} + \mu^{2}_{y} + c_{1}) (\sigma^{2}_{x} + \sigma^{2}_{y} + c_{2})}
    \label{eq:ssim}
\end{align}

Where $c_{1} = (0.01 L)^{2} $ and $c_{2} = (0.03 L)^{2}$, and $L$ is defined by the dynamic range of the pixel values. 
% We evaluate classification performance using accuracy, precision, and recall.

\subsection{VGG-16 on ImageNet}
As a second experiment, we apply our loss function to VGG-16~\cite{DBLP:journals/corr/SimonyanZ14a}, trained on the ImageNet dataset~\cite{DBLP:journals/ijcv/RussakovskyDSKS15} for an image classification task. Again we train two trustworthy models, one using Grad-CAM to generate saliency maps during training, the other using Guided Grad-CAM. We demonstrate improved post hoc saliency maps as evaluated by users, and improved classification performance. We compare this to a VGG-16 baseline trained using only the cross entropy loss. We use a subset of $30,000$ images, and update the parameters of VGG-16 from the $block5\_conv3$ layer. This layer was chosen as it is the closest convolutional layer to the softmax layer. We freeze the weights of all other layers. We compare classification performance of all models using accuracy, precision, and recall. We train different models varying the learning rate and lambda $\lambda$. We perform a grid search for the learning rate and lambda hyperparameters. We consider learning rates $\in \{.01, .1, .3\}$, and $\lambda \in \{ .9, .7, .5\}$. Post hoc saliency maps are evaluated with a user experiment, detailed below.
%We train with the proposed loss to find new parameters for this layer that are saliency map aware. 
\subsubsection{User Experiment}

To compare the post hoc saliency maps across models, a scoring metric is needed. Two commonly used metrics are localization error~\cite{DBLP:journals/corr/SelvarajuDVCPB16}, or the pointing game~\citep{DBLP:journals/corr/SelvarajuDVCPB16, DBLP:conf/eccv/ZhangLBSS16}. Both methods require ground truth object labels. We do not assume the data has any object annotations, hence these metrics cannot be used.

To score the post hoc saliency maps between the trustworthy gradient penalized models and their respective baselines, we conduct a user experiment. We take the best performing set of hyperparameters (learning rate of $.01$, and $\lambda=.9$), and use all models in the experiment to generate saliency maps for user evaluation.
% We denote these model Grad-CAM VGG-16, and Guided Grad-CAM VGG-16, respectively.

We randomly sample test set images for user evaluation. We show users the input image and a post hoc saliency map from each model. We ask $29$ users "Which image best highlights the <true\_class> in the original image." Users were given the option of "Can't distinguish" and "They look the same" in the case that no saliency maps are convincing. Users are asked if they have a Bachelor's degree in Computer Science, if they are familiar with the term saliency map used by the machine learning community, and if they have experience in computer vision. A link to the experiment can be found below.~\footnote{https://forms.gle/DMszuv84sbxB9tzt7}

\section{Results}
\subsection{Transfer Learning}

The classification performance for all models can be found in Table~\ref{fig:transfer-learning-table}. We observe the trustworthy MobilenetV2 model trained using Guided Grad-CAM outperformed the baseline and trustworthy Grad-CAM model. We find that both trustworthy CNN models outperformed their respective baseline models trained with just the cross entropy term.

The SSIM between saliency maps of the baseline model and trustworthy CNN trained with Grad-CAM was $.78$, and $.64$ between the baseline and trustworthy CNN trained with Guided Grad-CAM. When $R_{1} = 0$, the SSIM for the trustworthy Grad-CAM and Guided Grad-CAM models drop to $.76$, and $.5$ respectively. When $R_{2}=0$, the SSIM for the trustworthy Grad-CAM and Guided Grad-CAM models increases to $.85$, and $.94$ respectively. This metric shows the post hoc saliency maps differ visually from the baseline, however, it fails to identify which saliency maps are more visually correct.

%Figure~\ref{fig:cats-dogs-maps} shows post hoc saliency maps generated on several examples.

\begin{table}[t]
  \caption{Classification Performance-MobilenetV2}
  \centering
  \begin{tabular}{llll}
    \toprule
    Model&Accuracy&Precision&Recall \\
    \midrule
    MobilenetV2 Baseline&97.6\%&97.6\%&97.6\%  \\
    Trustworthy CNN w/ Grad-CAM&98.4\%&98.4\%&98.4\%  \\
    Trustworthy CNN w/ Grad-CAM ($R_{1}=0$) &98.3\%&98.3\%&98.3\%\\
    Trustworthy CNN w/ Grad-CAM ($R_{2}=0$) &98.5\%&98.5\%&98.5\%\\
    \textbf{Trustworthy CNN w/ Guided Grad-CAM}&\textbf{98.7\%}&\textbf{98.7\%}&\textbf{98.7\%} \\
    Trustworthy CNN w/ Guided Grad-CAM ($R_{1}=0$) &98.6\%&98.6\%&98.6\%\\
    Trustworthy CNN w/ Guided Grad-CAM ($R_{2}=0$) &98.4\%&98.4\%&98.4\%\\
    \bottomrule
  \end{tabular}
  \label{fig:transfer-learning-table}
  \caption*{Classification performance on transfer learning task. $R_{1}=0$ denotes models trained with equation~\ref{eq:R1} set to zero, $R_{2}=0$ denotes models trained with equation~\ref{eq:R2}.}
\end{table}

\begin{table}[t]
  \caption{Classification Performance-VGG16}
  \centering
  \begin{tabular}{llll}
    \toprule
    %\multicolumn{2}{c}{Part}                   \\
    \cmidrule(r){1-4}
    %Learning rate &.01&.01&.01&.1&.1&.1&.3&.3&.3 \\
    %\cmidrule(r){1-10}
    &Accuracy & Precision & Recall \\
    \midrule
    VGG-16 Baseline &66\% (56\%,75\%)&49.7\%&50.8\%\\
    Trustworthy CNN w/ Grad-CAM &\textbf{70\% (61\%, 78\%)}&\textbf{54.6\%}&\textbf{55.7\%}\\
    \cmidrule(r){1-4}
    VGG-16 Baseline &66\% (56\%,75\%) &49.7\%&50.8\%\\
    Trustworthy CNN w/ Guided Grad-CAM &\textbf{70\% (61\%, 78\%)}&\textbf{54.6\%}&\textbf{55.7\%}\\
    \bottomrule
  \end{tabular}
  \label{fig:vgg-performance}
  \caption*{Classification performance shown for gradient penalized trustworthy models and baselines on test set images shown during user experiment. A $95\%$ confidence interval (lower, upper) is also included. }
\end{table}

\subsection{VGG-16 on ImageNet}

Table~\ref{fig:vgg-performance} shows the accuracy, precision, and recall of all models on the subset of test set images shown to users in the experiment. We use the best performing set of hyperparameters to be evaluated by the users. We find that both trustworthy models outperform their respective baselines. We recognize the classification performance between the trustworthy models to be equal, likely due to setting $\lambda=.9$. 

%We find the best performing gradient penalized models had a learning rate of $.01$ and $\lambda = .9$. 

\subsubsection{User Experiment}

We find the SSIM between the baseline VGG-16 and trustworthy Grad-CAM model was $99.8$, and $99.4$ between the baseline VGG-16 and trustworthy Guided Grad-CAM model. According to this metric, the saliency maps generated by the gradient penalized models should be very similar to the base model. Through our user experiment however, we find this not to be the case. This is further discussed in Section~\ref{limitations}.

Table~\ref{fig:user-experiment} further breaks down the user experiment, showing the percentage of images that fall into each case. Users decided both trustworthy models outperform the baseline in all scenarios, except case $2$. Recall case $2$ occurs when the predicted label is incorrect and the saliency map is accurate. Ideally images fall into case $1$, and fewer images fall into cases $2$, $3$, and $4$.

\begin{table}[h]
  \caption{User Experiment Breakdown-VGG16}
  \centering
  \begin{tabular}{lll}
    \toprule
    \cmidrule(r){1-3}
    &Case $1$ &Case $2$\\
    \midrule
    VGG-16 Baseline &18\% (7\%,28\%)&\textbf{16\% (6\%, 26\%)}\\
    Trustworthy CNN w/ Grad-CAM &\textbf{44\% (30\%, 58\%)}&\textbf{16\% (6\%, 26\%)}\\
    \midrule 
    VGG-16 Baseline &16\% (6\%, 26\%)&\textbf{10\% (2\%,18\%)}\\
    Trustworthy CNN w/ Guided Grad-CAM &\textbf{50\% (36\%, 64\%)}&18\% (7\%, 29\%)\\
    \bottomrule
    \bottomrule
        \cmidrule(r){1-3}
    & Case $3$&Case $4$\\
    \midrule
    VGG-16 Baseline &44\% (30\%, 58\%)&16\% (6\%, 26\%)\\
    Trustworthy CNN w/ Grad-CAM &\textbf{22\% (10\%, 33\%)}&\textbf{12\% (3\%, 21\%)} \\
    \midrule
    VGG-16 Baseline &48\% (34\%, 62\%)&20\% (9\%,31\%)\\
    Trustworthy CNN w/ Guided Grad-CAM &\textbf{18\% (7\%, 29\%)}&\textbf{8\% (0\%, 20\%)}\\
  \end{tabular}
  
  \label{fig:user-experiment}
  \caption*{For the Trustworthy CNN trained with Grad-CAM, $44\%$ of images shown to users were found to have accurate predictions, and more accurate saliency maps (relative to the baseline). A $95\%$ confidence interval (lower, upper) is included. Recall Case $1$: \% of observations with predicted labels correct and resulting saliency maps accurate. Case $2$: \% of observations with predicted labels incorrect and resulting saliency maps accurate. Case $3$: \% of observations with predicted labels correct and resulting saliency maps inaccurate. Case $4$ \% of observations with predicted labels incorrect and resulting saliency maps inaccurate. High percentage is desirable for Case $1$, low percentage is desirable for Cases $2$, $3$, and $4$.} 
\end{table}

\subsection{Discussion}
We find the trustworthy models trained with Guided Grad-CAM outperform all other models in terms of predicting correct labels and accurate post hoc saliency maps. In the transfer learning experiment, we find the SSIM decreases on gradient penalized models when equation~\ref{eq:R1} is set to zero. Additionally, the SSIM increases when equation~\ref{eq:R2} is set to zero. This shows the inherent trade-off between the two terms.

On the ImageNet dataset, users found the baseline VGG-16 models to produce inaccurate saliency maps. These models are not trustworthy. The most common error from the baseline models was producing accurate predictions with inaccurate saliency maps (case $3$ from Table~\ref{fig:user-experiment}). This is not surprising considering the cross entropy loss is saliency map unaware. Users will question the legitimacy of the model when inaccurate saliency maps are produced. Our approach offers improved classifiation results and more accurate saliency maps, resulting in increased user trust.

%User trust and model performance can be improved using the saliency map aware loss function from equation~\ref{eq:loss}. 

\section{Limitations}
\label{limitations}
\subsection{Loss Function}
One noticeable limitation using the proposed loss is that only one convolutional layer can be updated at a time in training. This is partially due to the limitations of some saliency methods. Grad-CAM and Guided Grad-CAM~\cite{DBLP:journals/corr/SelvarajuDVCPB16} for example generate saliency maps using the gradients from a specific layer only. Hence the gradients of one individual layer are used to compute the saliency map. This layer however may not fully represent what the entire model has learned. %It is frequently debated whether post hoc saliency maps are the best approach for model interpretation. 
\begin{figure*}[h]
	\centering
	\begin{subfigure}{0.24\textwidth}
		\includegraphics[width=\textwidth]{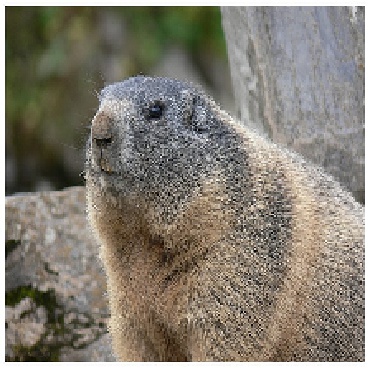}
		\caption{Input Image}
		\label{fig:animal-base}
	\end{subfigure}
	\begin{subfigure}{0.24\textwidth}
		\includegraphics[width=\textwidth]{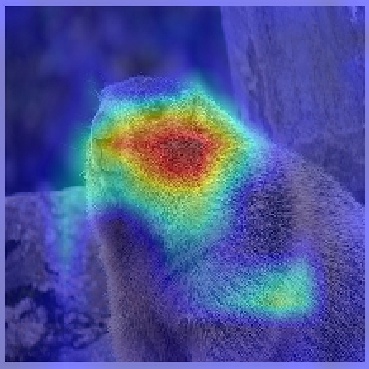}
		\caption{Hypothetical Model 1}
		\label{fig:animal-gc-one}
	\end{subfigure}
	\begin{subfigure}{0.24\textwidth}
		\includegraphics[width=\textwidth]{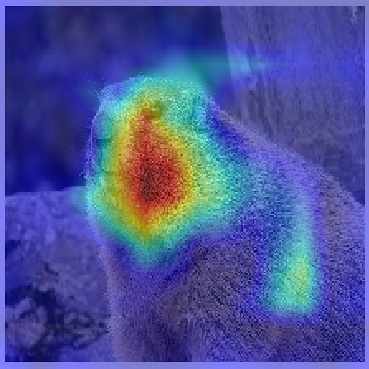}
		\caption{Hypothetical Model 2}
		\label{fig:animal-gc-two}
	\end{subfigure}
	\caption{Two saliency maps with equal attribution values}
	\label{fig:animal-fig}
\end{figure*}

\subsection{Saliency Map Scoring}
%Assigning a score to the saliency map based on how well it attributes pixels over the object of interest can be problematic. 

One difficulty in scoring saliency maps is that two saliency maps can correctly highlight the object of interest, but equal attribution values can be assigned to different parts of the same object. An example of this can be demonstrated in Figure~\ref{fig:animal-fig} using two hypothetical models. 

For some input image, both models output saliency maps with equal total attribution values, but pixels are attributed to different locations on the object. The model in Figure~\ref{fig:animal-gc-one} attributes the face of the Marmot, and the model in Figure~\ref{fig:animal-gc-two} attributes a portion of the face and body. These two saliency maps have exactly the same total attribution values when averaged. Hence, the SSIM between Figures~\ref{fig:animal-gc-one} and~\ref{fig:animal-gc-two} is $99.9$, but look significantly different. It is unclear which saliency map is more visually accurate.

%One solution to this is to let users determine. Evaluating the faithfulness of post hoc saliency maps through user experiments however is not always feasible for large datasets. 

\section{Conclusion}
In this work, we combine the use of post hoc interpretability methods with traditional loss functions to learn trustworthy model parameters. We propose a loss function that penalizes inaccurate saliency maps during training. Further constraining the loss function used by convolutional neural networks increases classification performance, and users found the post hoc saliency maps to be more accurate. This give a more dependable model. Future work involves extending this method to other tasks (image captioning, object tracking, etc), and other deep learning architectures. 

\section*{Broader Impact}
%a) who may benefit from this research, b) who may be put at disadvantage from this research, c) what are the consequences of failure of the system, and d) whether the task/method leverages biases in the data. If authors believe this is not applicable to them, authors can simply state this.
Users receiving an automated decision from a convolutional neural network will benefit from this research; Our approach provides a way to increase user trust in models previously treated as black box.

Using this approach, parameters of a pre-trained model can be updated, or parameters of a new layer can be learned in a transfer learning framework. Errors from an existing model can be identified and fixed. For practitioners wanting to eliminate a race or gender bias from a model, they will not have to retrain the model from scratch. This will save electricity used by the machine(s) to train.  

We do not believe anyone is put at a disadvantage from this research. A failure of this system would mean the model would no longer be convincing to users, and thus no different than the original black box model. 

\bibliographystyle{plain}
\bibliography{bibfile}

\end{document}